\documentclass[twocolumn]{article}

\usepackage{microtype}
\usepackage{graphicx}
\usepackage{subfigure}
\usepackage{booktabs} 
\usepackage{authblk}
\graphicspath{{icml2019_style/}}
\usepackage{hyperref}

\begin{document}
\title{Robust or Private? Adversarial Training Makes Models More Vulnerable to Privacy Attacks}




\author[1]{Felipe A. Mejia}
\author[2]{Paul Gamble}
\author[1]{Zigfried Hampel-Arias}
\author[1]{Michael Lomnitz}
\author[1]{Nina Lopatina}
\author[1]{Lucas Tindall}
\author[2]{Maria Alejandra Barrios}

\affil[1]{Lab41, In-Q-Tel, Menlo Park, USA}
\affil[2]{Previously at Lab41, In-Q-Tel, Menlo Park, USA}



\maketitle
\vskip 0.3in




\begin{abstract}
Adversarial training was introduced as a way to improve the robustness of deep learning models to adversarial attacks. This training method improves robustness against adversarial attacks, but increases the model’s vulnerability to privacy attacks. In this work we demonstrate how model inversion attacks, extracting training data directly from the model, previously thought to be intractable become feasible when attacking a robustly trained model. The input space for a traditionally trained model is dominated by adversarial examples - data points that strongly activate a certain class but lack semantic meaning - this makes it difficult to successfully conduct model inversion attacks. We demonstrate this effect using the CIFAR-10 dataset under three different model inversion attacks, a vanilla gradient descent method, gradient based method at different scales, and a generative adversarial network base attacks. 

\end{abstract}

\section{Introduction}
\label{introduction}

Machine learning models have gathered a large amount of success in recent years, with applications in several industries. These models are increasingly being used in safety critical scenarios with data that is both public or private, such as health data or customer photos. As we increasingly rely on machine learning, it is critical to assess the dangers of the vulnerabilities in these models. 

In a supervised learning environment several components make up the learning pipeline of a machine learning model: collection of training data, definition of model architecture, model training, and model outputs -- test and evaluation. This pipeline presents a broader attack surface, where adversaries can leverage system vulnerabilities to compromise data privacy and/or model performance. There are four main types of attacks discussed in literature: 
\begin{itemize}
  \item model fooling -- small perturbation to the user input leads to large changes in the model output \cite{goodfellowexplaining}, \cite{carlini2017adversarial}
  \item model extraction -- the model is reverse engineered from its outputs \cite{tramer2016stealing}.
  \item model poisoning-- the training process is altered to create vulnerabilities at inference \cite{hitaj2017deep}, \cite{yang2017generative}.
  \item model inversion-- the model is used to re-generate or gain information on the training data \cite{yang2017generative}, \cite{shokri2017membership}
\end{itemize}
However these attacks are assumed to be performed in vacuum, with little discussion on the interplay between defense strategies for different types of attacks. In this paper we focus on model fooling and model inversion, with emphasis on how mitigating one attack can increase vulnerabilities in the other. 
\begin{figure}[ht]
\vskip 0.2in
\begin{center}
\centerline{\includegraphics[width=\columnwidth]{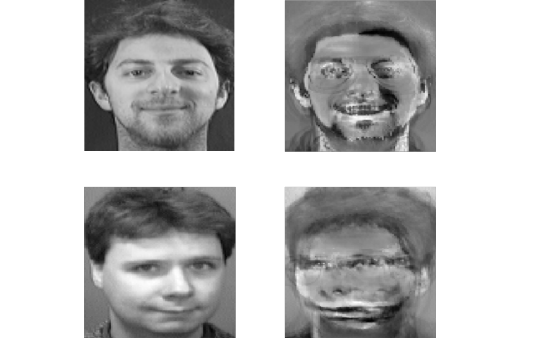}}

\caption{Model inversion results on ATT data. }
\label{ATT}
\end{center}
\vskip -0.2in
\end{figure}

Model inversion attacks directly attack the privacy of the training dataset. There are several scenarios where the data needs to be kept private either for financial reasons or because we want to protect the training members privacy. For example if we train our model on text data that includes social security numbers, we do not want anyone to be able to access such information from the training data if we make the model public. Previous work on model inversion has explored small datasets and small models. \cite{fredrikson2015model} performed model inversion of facial images on a logisitic regression model trained with the ATT dataset \cite{att_faces}. Although the methods were successful on the small dataset (400 gray-scale images) they do not directly extend to larger datasets as demonstrated by \cite{shokri2017membership}. Even with the small dataset and model, this approach only works on classes that do not vary in pose or position within the dataset, as demonstrated in Figure \ref{ATT}.  Here the top images demonstrate a successful reconstruction and the bottom images show results, using the same methodology, on another category that is not well reconstructed. \cite{shokri2017membership} demonstrated the difficulty of model inversion by applying the Frederickson attack to a convolutional neural network trained on the CIFAR-10 dataset. The reconstructed images demonstrated no recognizable semantic meaning demonstrating the ineffectiveness of the attack. 

Model fooling attacks have gathered much more attention and success. With several variations of a gradient based attack, the model’s output can be drastically changed with a small change in the input to the model. Several different defenses have been proposed to alleviate this problem, including model distillation, image blurring and denoising \cite{papernot2016distillation}, \cite{liao2018defense}, \cite{carlini2017adversarial}. Recently, a promising defense has been obtained with the idea of adversarially trained models (ATM) \cite{madry2017towards}, \cite{tsipras2018robustness}. This involves formulating the traditional training process as a min-max problem. First an adversarial example is maximized to fool the model, while staying within a L2 distance from the training image. Then the model is trained to minimize the loss to these adversarial examples.  \cite{madry2017towards} demonstrated that adversarial training also comes with other consequences, namely a trade-off with accuracy as well as better alignment of semantic representation between a model and human perception. 

In this paper we explore how this improved semantic representation of ATM affects the privacy of training data specific to model inversion attacks. We demonstrate that by improving the semantic representation that models have, we are able to generate better model inversion reconstructions.

\section{Approach}
\label{Approach}
This work focuses on three image classification models, referred here as the target models. The target models are trained to classify images of the CIFAR-10 dataset \cite{krizhevsky2009learning}, \cite{krizhevsky2010convolutional}. One model is robustly trained and the other two models are trained in a traditional manner. The target models are then attacked with three different model inversion attacks. In this scenario the goal of the model inversion attack is to reconstruct an image from the training dataset.
\subsection{Target Models}
The first model that we test is a traditionally trained model(TTM) using the VGG16 architecture (TTM-VGG16) changing the fully connected layers to handle the CIFAR image size \cite{simonyan2014very}. The model was trained for 100 epochs with a batch size of 128, using an Adam optimizer with a learning rate of 0.001. This resulted in a model with 82.2\% validation accuracy and 99.2\% training accuracy. 

The second model that was used is based on the w28-10 wide-resnet architecture and again was traditionally trained (TTM-Res) \cite{zagoruyko2016wide}. The model optimizer was stochastic gradient descent with momentum of 0.9 and weight decay of 0.0002. The learning rate was 0.01 for 100 epochs. The final accuracies were 84.9\% and 95.7\%  for the validation and training respectively.

The final model, an adversarially trained model (ATM), is based on the w28-10 wide-resnet architecture (ATM-Res) following the work of \cite{madry2017towards}. This model was trained with adversarial examples generated with the Fast gradient sign method (FGSM) \cite{goodfellowexplaining}. The attack was allowed to perturb the image up to 10 pixel counts with the image range being from 0-255. The step size of the attack was 2 and the attack had 10 iterations to find an adversarial example. The model optimizer was stochastic gradient descent with momentum of 0.9 and weight decay of 0.0002. The learning rate was 0.1 for the first 100 epochs and 0.01 for the next 50 and 0.001 for the following 50 epochs, for a total of 200 epochs. The final accuracies were 78.5\% and 99.7\%  for the validation and training respectively. 

\subsection{Model Inversion Attacks}
The first attack that we explore is a basic gradient based attack. Here we start by inputting a blank gray image (128 for 0-255 range) and backpropagate to optimize the image such that it maximizes a particular output category that we are interested in extracting. We start with a blank image instead of random inputs as the optimization does not remove all of the randomness that it is initialized with and a crisper reconstruction is generated with a homogenous image input. The image does not necessarily have to start as gray and other homogeneous initialization could be used but gray gives best results for the most classes. Instead of changing the weights of our model, as in regular training, we change the input pixels of the image to maximize a category that we want to extract. This method is very similar to the projected gradient descent (PGD) that is used for model fooling, mainly changing the input from a natural image to a blank input. The gradient step is calculated as follows,

\begin{equation}
\label{PGD_step}
    X_{t+1} = clip(X_t + lr * G)
\end{equation}

with $X_{t}$ being the image at iteration t, clipping is applied to keep the image within the 0-255 range. G is the loss gradient with respect to the image, where the loss is the output of the model for a particular class before softmax.

The next model inversion attack is based on the principle of Google’s DeepDream method of optimizing at different image resolution scales \cite{mordvintsev2015deepdream}. The way the original DeepDream algorithm works is by taking a natural image as an input and scaling it down a preset number of octaves with each octave scaling down the image by a preset octave scale. This scaled down image is then scaled back up to the original image dimensions effectively blurring the image and focusing on low frequency features. Once this low resolution image has been optimized it is upscaled and reoptimized at a higher resolution until all octaves are optimized. We use this technique for model inversion by inputting a blank image and running through the entire DeepDream method for 5 iterations. We use a downsampling octave scale of 2 and run a total of 4 octaves for each downsampled image we run 10 iterations of gradient descent optimization. We also added normalization loss as in \cite{mahendran2016visualizing}. The gradient step is modified from the PGD attack as,  

\begin{equation}
\label{Dream_step}
    X_{t+1} = clip\left(X_t + lr * \frac{G}{|G|_{mean}}\right)
\end{equation}
where  $\frac{G}{|G|_{mean}}$ is the average of the absolute value of the gradients at that iteration step.

\begin{figure}[ht]
\vskip 0.2in
\begin{center}
\centerline{\includegraphics[width=\columnwidth]{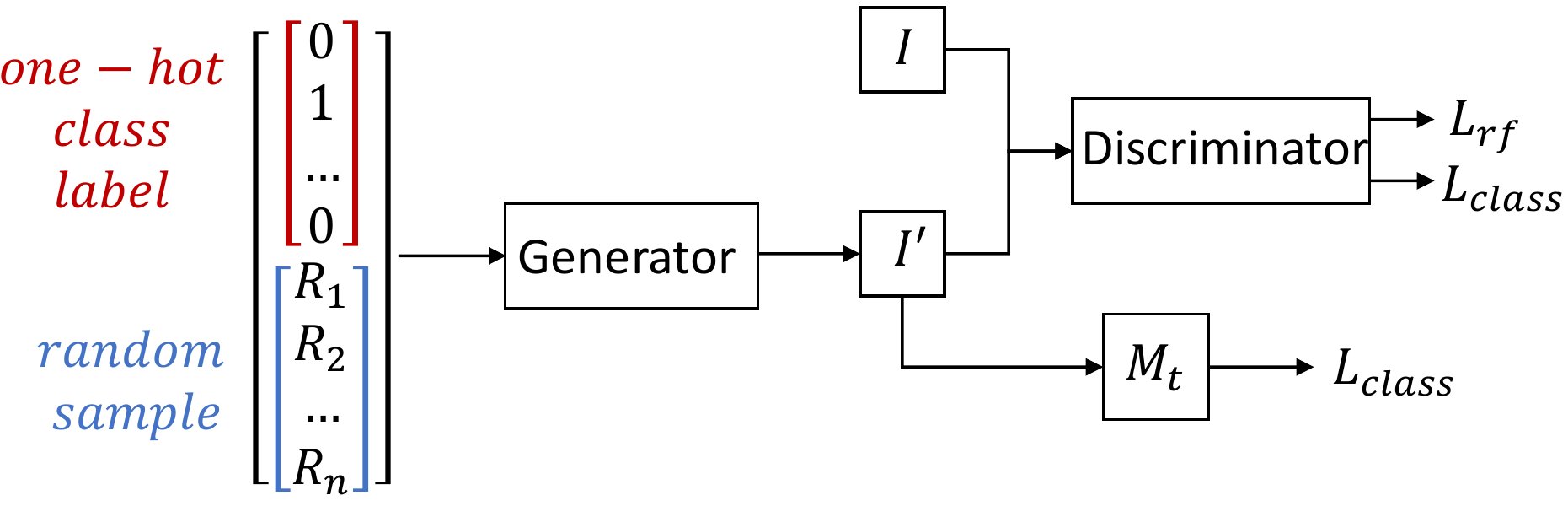}}
\caption{Diagram of the generative adversarial network used for model inversion. a vector of one-hot class labels and random normal numbers is  input to the generator. The generator creates an image I' which is then fed to the discriminator with real images. The discriminator is trained to classify real images and to discriminate real images from generated images. The generated images are also fed to the target model and the generator is trained to generate real images and fool the target model. }
\label{GAN_figure}
\end{center}
\vskip -0.2in
\end{figure}

\begin{figure*}[t!]
\vskip 0.2in
\begin{center}
\centerline{\includegraphics[width=\textwidth]{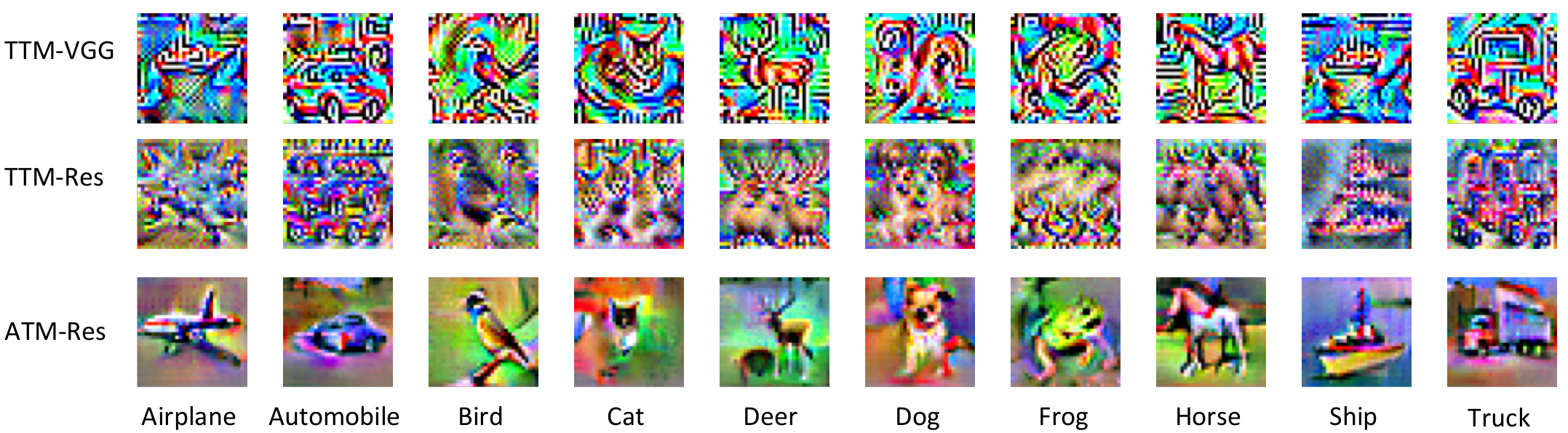}}
\caption{Reconstructed images of the 10 CIFAR-10 classes generated from gradient ascent model inversion attack. Top middle and bottom row are attacking the TTM-VGG, TTM-Res and ATM-Res respectively.}
\label{CIFAR10_PGD_figure}
\end{center}
\vskip -0.2in
\end{figure*}

Figure \ref{GAN_figure} illustrates the architecture of the generative adversarial network (GAN) used for model inversion. The main goal of using a GAN for model inversion is to better constrain reconstructed images to be realistic images. We followed the auxiliary classifier (AC) GAN method, modifying it to use the classification loss from the target model \cite{odena2017conditional}. The method starts by inputting a vector of one-hot class labels followed by random numbers. This input vector is fed into the generator, which is composed of 7 transpose convolutions each followed by a ReLU. The generator outputs an image that is fed to the discriminator. The discriminator is made of 7 blocks of convolutional, batchnorm, leaky ReLUs, and dropout layers. The convolutional layers are 5 by 5 filters, the leaky ReLU has a negative slope of 0.2, and the dropout rate was 0.5. The discriminator is trained on a real-fake loss and a class loss, the class loss is only calculated for real images since the target model will classify the generated images. The real images that are used to train the discriminator are the validation images for the CIFAR-10 dataset representing what could be used in a real world attack to constrain the generated images to realistic images, this out of set images for model inversion is called shadow data in literature \cite{salem2018ml}. The generated images are also fed into the target model to calculate the class loss for the generator. The generator is trained on the discriminators real-fake loss as well as the target models class loss. The optimizer for both the discriminator and generator are Adam optimizers with a learning rate of 0.0002 and betas of 0.5 and 0.999. 

\section{Results}
\label{Results}

In this section we demonstrate the differences between attacking a TTM and an ATM. The main goal of this comparison is to understand how adversarial attacks and privacy attacks are related. We focus on answering three fundamental questions:

\begin{enumerate}
  \item Is there a difference in the vulnerability to privacy attacks by defending against adversarial attacks?
  \item What fundamentally is causing this relationship?
  \item Are privacy concerns enough to keep models from being robustly trained?
\end{enumerate}

\subsection{PGD model inversion}

Just like\cite{shokri2017membership} we observe that PGD attacks for model inversion do not reconstruct semantically meaningful images for TTM. At best the images generated can aide in identifying the label of a particular class, but does not provide any information on a particular data point. Images generated from TTM-VGG have very sharp edges and lines throughout with some slight semantic meaning. TTM-Res has slightly more semantic meaning but is plagued by fractal patterns of the category being generated (This is highlighted in \ref{DeepDream_subsection}). These images also have extensive checker patterns throughout the images. As shown in Figure \ref{CIFAR10_PGD_figure}, once a model is adversarially trained, the reconstructions become more semantically meaningful.  The images are clearer and focused on individual samples of the category.
\begin{figure}[ht]
\vskip 0.2in
\begin{center}
\centerline{\includegraphics[width=\columnwidth]{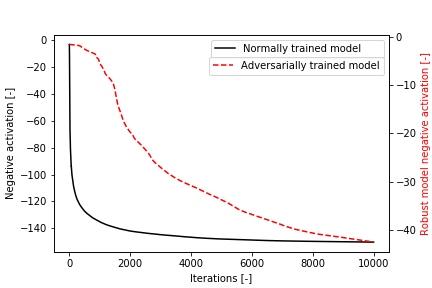}}
\caption{ The activation loss during the gradient descent process for 10000 iterations for generating a bird. The dashed and solid line represent the adversarially trained and traditionally trained model respectively. The y-axis on the left is for the normally trained model while the y-axis on the right is for the adversarially trained model. Note the order of magnitude difference in the two axis.}
\label{PGD_iterations}
\end{center}
\vskip -0.2in
\end{figure}

The optimization process of the reconstructions also significantly changes. As seen in Figure \ref{PGD_iterations} the TTM-Res quickly optimizes the image, creating an image that is classified as a bird within 3 iterations, while the ATM-Res takes 1136 iterations to be classified as the target class of bird. This highlights the prevalence of adversarial examples in the input space of the TTM. Even if we randomly initialize the images with different inputs, the TTM takes two orders of magnitude fewer iterations to generate an image that is classified as the class that we are targeting. Because the model embedding of the TTM does not have a strong semantic representation, a wide range of adversarial examples satisfy the target category boundary conditions-- for almost any image initialization there is an adversarial example that is near that input. 
\begin{figure}[ht]
\vskip 0.2in
\begin{center}
\centerline{\includegraphics[width=\columnwidth]{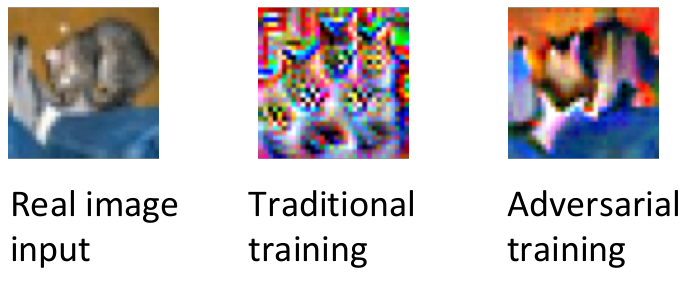}}
\caption{Reconstructed images from the input cat for traditionally and adversarially trained models.}
\label{ICML_cat}
\end{center}
\vskip -0.2in
\end{figure}

\label{DeepDream_subsection}

\begin{figure*}[t!]
\vskip 0.2in
\begin{center}
\centerline{\includegraphics[width=\textwidth]{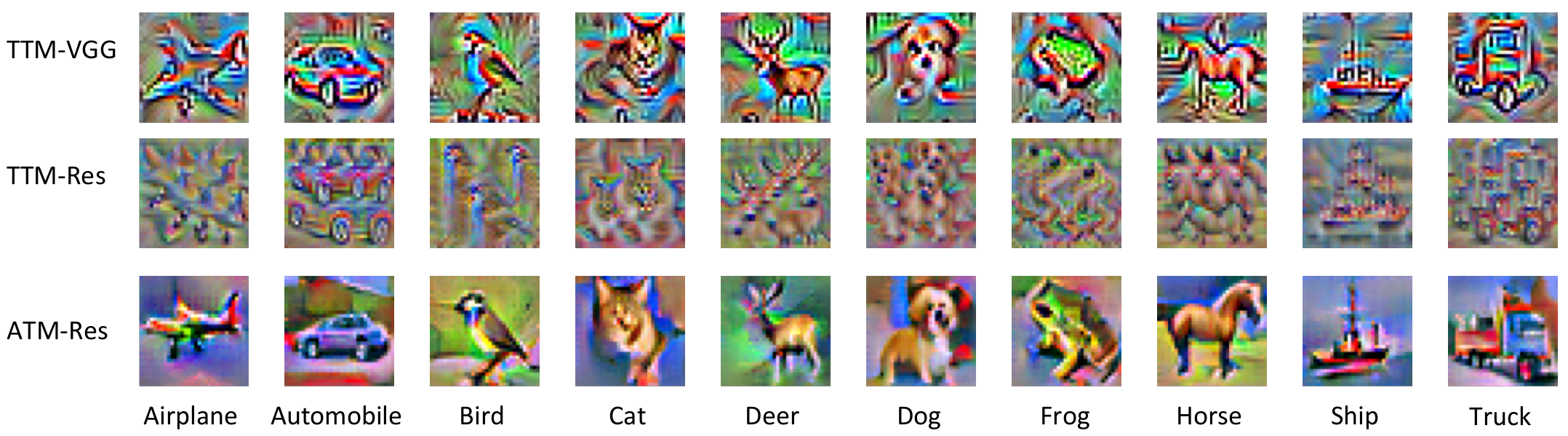}}
\caption{Reconstructed images of the 10 CIFAR-10 classes generated from deep dream model inversion attack. Top middle and bottom row are attacking the TTM-VGG, TTM-Res and ATM-Res respectively.}
\label{DeepDream_CIFAR}
\end{center}
\vskip -0.2in
\end{figure*}

The activation values of TTM-Res, and ATM-Res differ by almost an order of magnitude, with the bird activations being 150 and 44 for the TTM-Res and ATM-Res respectively. The average activation value on the training data is 11 and 7 for the traditionally trained model and adversarially trained model respectively. When we input the reconstructed image from the ATM-Res into the TTM-Res we get a bird activation value of 21 still significantly smaller than 150, but much higher than the average training image. This demonstrates that all of the reconstructions have higher activations than naturally occurring images. The adversarially trained reconstruction is not a minimum in the TTM-Res because of the nearby adversarial examples and once those examples are removed it becomes a minimum.  

Inputting a real training image into the model inversion we can further distinguish the characteristics of the two models. In Figure \ref{ICML_cat} we can see that inputting a cat image from the training data to TTM-Res, the model inversion attack begins to generate multiple extra cats and drastically changes the image. The overall image information is lost and none of the shapes are preserved. While for ATM-Res, the reconstructed image focuses more on emphasizing traits that defined a cat in the entire dataset and focuses on contrasting the cat from the background. The robustly trained model reconstruction also begins to remove information about the background, for example blurring the foot and pants. 
 
\subsection{DeepDream model inversion}
DeepDream model inversion is able to push the image reconstruction away from adversarial examples through its use of multiple scales. This acts as a filter focusing the reconstruction first on low frequency features and then adding higher frequency characteristics as the reconstruction is scaled up. This greatly improves the reconstruction in TTM as demonstrated in Figure \ref{DeepDream_CIFAR}. But since ATM have a decreased prevalence of adversarial examples this method does not improve the model inversion on these models. This method is more successful with TTM-VGG than TTM-Res. TTM-Res tend to generate reconstructions that attempt to have multiples of the class in question. As seen Figure \ref{DeepDream_CIFAR} the TTM-Res generated images have several wheels, antlers, cats, dogs, for the classes of trucks, deer, cats and dogs respectively. This is one of the defining traits in the DeepDream method, but is not observed in the TTM-VGG, suggesting that the skip connections in the resnet architecture allow the model to consider more examples of a class as more likely being that class. This characteristic make model inversion more difficult as small scale features can be extracted but larger scale features are more difficult to obtain. By adversarially training the resnet model, this characteristic of having multiple examples of the class in one image is also removed. The ATM-Res extracted images are highly focused on one single, specific, individual and create distinct contrast between the main subject/category and the image background. 

The background of the extracted images highlights what remains private in the model, with backgrounds of dogs mainly being contrasting colors, while the background for planes, deer, horses, boats, trucks have blue skies, grass, open field, water, and roads reflecting backgrounds consistent within all of these classes while dogs have more varied backgrounds in the dataset. 

\subsection{GAN model inversion}
By using a GAN we are further constraining the image generation process to generate real images. The goal is to  push the generated images away from adversarial images or non-semantically meaningful images, toward realistic looking images. For the TTM the training process quickly finds a way to fool the target network, in essence adding a small amount of noise and focusing on fooling the discriminator. This limits the ability to extract information from the target model as the generated images are more representative of the shadow dataset than the target dataset.

\begin{figure*}[t!]
\vskip 0.2in
\begin{center}
\centerline{\includegraphics[width=0.9 \textwidth]{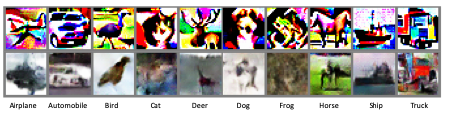}}
\caption{Reconstructed images of the 10 CIFAR-10 classes generated from a GAN model inversion attack with different weights on the target class.}
\label{GAN_CIFAR}
\end{center}
\vskip -0.2in
\end{figure*}

\section{Discussion}
\label{Discussion}

\subsection{Generating different samples}
\begin{figure}[ht]
\vskip 0.2in
\begin{center}
\centerline{\includegraphics[width=\columnwidth]{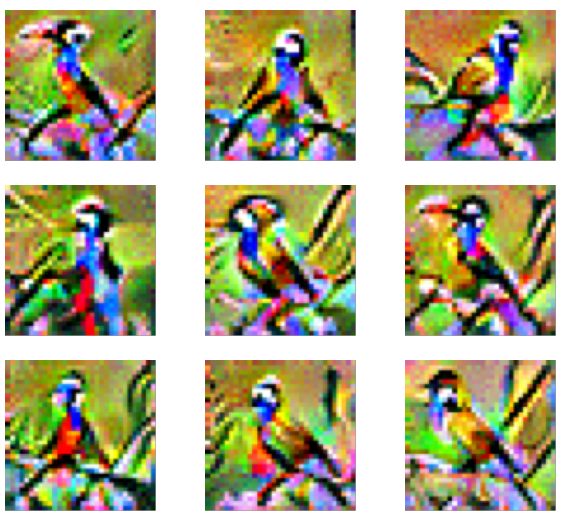}}
\caption{Nine different example birds reconstructed with PGD model inversion using random inputs.}
\label{Bird_figure_PGD}
\end{center}
\vskip -0.2in
\end{figure}

Although the number of training images that are vulnerable to model inversion are limited, it is possible to obtain different reconstructions using different inputs. For PGD model inversion, we can input homogeneous images but that limits our reconstruction to a handful of images. We can also input random values but this hurts the overall structure of the image. As seen in Figure \ref{Bird_figure_PGD} the reconstructions do generate various birds but start loosing the detailed information. Furthermore to create more varied birds one would require larger variance in the noise, which further amplifies the loss in semantic representation of the images. Blurring and incorporating an L2 loss to the optimization can aid with these problems but does not improve significantly. 

\begin{figure}[ht]
\vskip 0.2in
\begin{center}
\centerline{\includegraphics[width=\columnwidth]{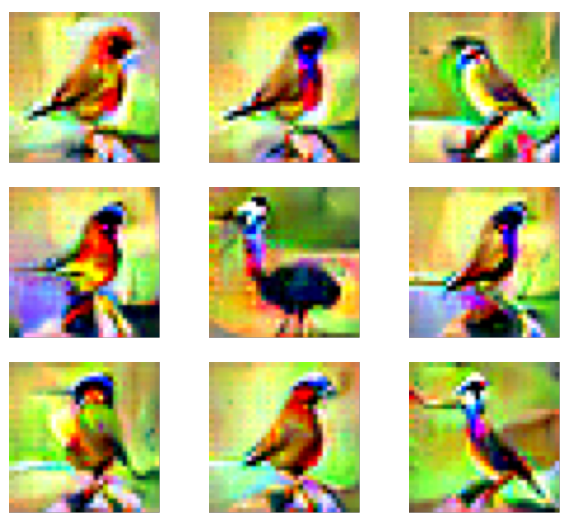}}
\caption{Nine different example birds reconstructed with DeepDream model inversion using random inputs.}
\label{Bird_figure}
\end{center}
\vskip -0.2in
\end{figure}

Figure \ref{Bird_figure} illustrates nine different birds obtained from the DeepDream model inversion. The DeepDream model inversion is more suitable for inputting different random inputs as it naturally blurs and smooths out the random input into the reconstructed image. In Figure \ref{Bird_figure} we can see that the color of the birds changes but also some of the shapes, namely an apparent ostrich and hummingbird in the middle and bottom left image respectively.

\subsection{Model inversion metrics}
\begin{figure*}[t!]
\vskip 0.2in
\begin{center}
\centerline{\includegraphics[width=\textwidth]{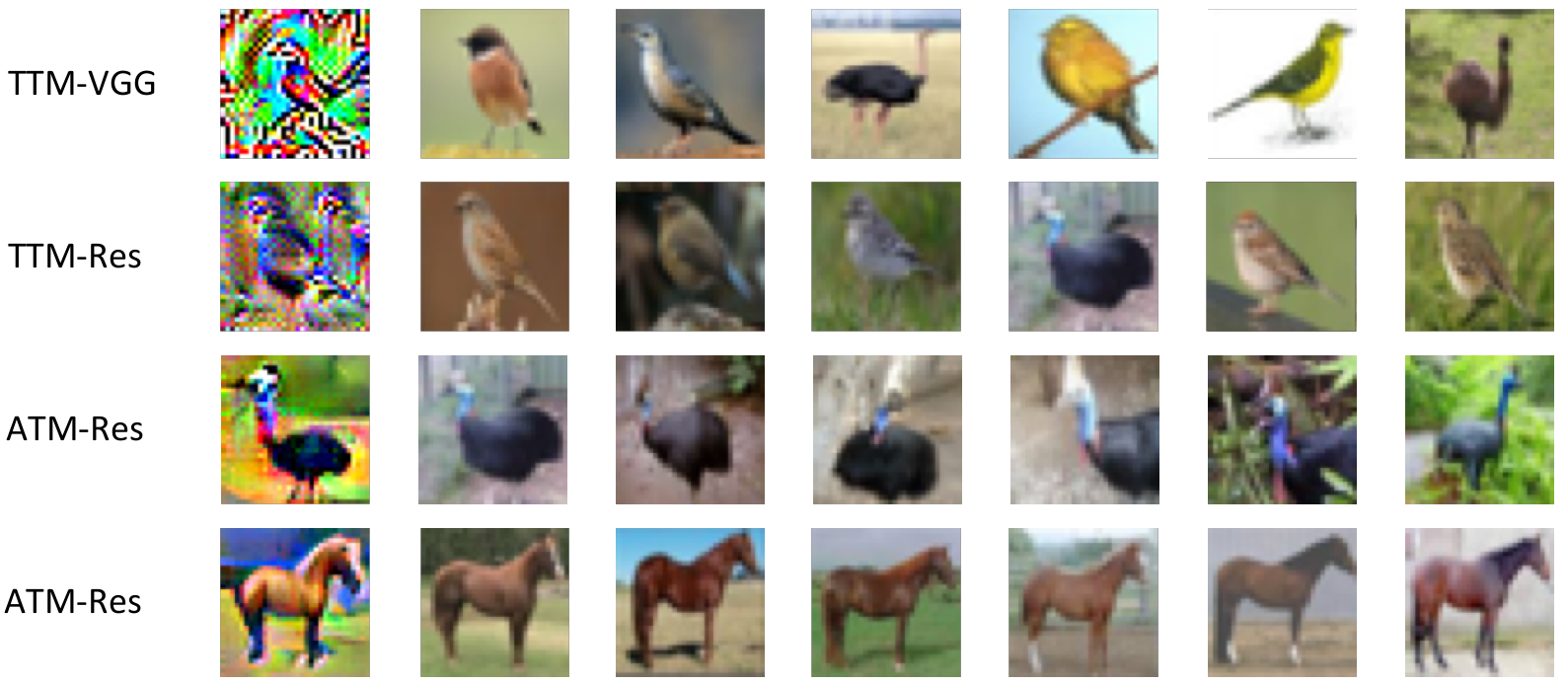}}
\caption{Nearest cosine similar training images to model inversion image on the left. First row is for TTM-VGG bird, second row is TTM-Res bird, third row is ATM-Res bird and fourth row is ATM-Res horse.}
\label{training_birds}
\end{center}
\vskip -0.2in
\end{figure*}

The success of the model inversion attack is one of the main challenges when developing different attack methods. When evaluating the attack there are two main challenges, finding the nearest image in the dataset and quantifying the similarity between the reconstructed image and the image in the dataset. For this reason previous model inversion attacks were analyzed qualitatively. 

We take a two prong approach to provide a quantitative metric to judge model inversion attack efficacy--  identifying the nearest data point to the reconstructed image and quantifying the similarity separately. First we identify the most similar data point by inputting the images into the target model, and extract a feature vector from the final convolutional layer. This feature vector is then used to find the cosine similarity of the nearest image to our reconstructed image. Figure \ref{training_birds} shows training images that were found to be the closest to our model inversion attack reconstruction (here for bird category). Note training images closest to our TTM model reconstructions have large variability in pose, type of bird, and background, whereas those found for the ATM reconstruction pertain to the same type of bird. Here we can see the blue neck and black body of the birds similar to our reconstruction. To show this is not category dependent, we also show results for horse for the ATM.   

\begin{figure}[ht]
\vskip 0.2in
\begin{center}
\centerline{\includegraphics[width=\columnwidth]{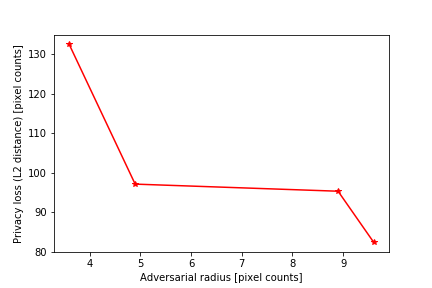}}
\caption{Adversarial radius vs privacy loss, for TTM-VGG, TTM-Res, ATM-Res10, and ATM-Res from left to right.}
\label{adversarial_privacy}
\end{center}
\vskip -0.2in
\end{figure}

ATM-Res and TTM-Res have an average maximum similarity score of 0.85 and 0.78 respectively while TTM-VGG has an average similarity score of 0.99. This suggest that TTM-VGG cannot differentiate intra-class variability. Once we obtain the image that the model believes is most similar we can then calculate the L2 image distance to quantify the similarity of our reconstructed image to the nearest training image. The average L2 is 82.4, 132.5, 97.1 for ATM-Res, TTM-VGG, and TTM-Res respectively. Figure \ref{adversarial_privacy} demonstrates the trade-off between the privacy loss we describe and an adversarial radius. The adversarial radius is calculated by applying the PGD adversarial attack until the image is misclassified. The four points on Figure \ref{adversarial_privacy} are from the TTM-VGG, TTM-Res, ATM-Res10, ATM-Res, where ATM-Res10 is ATM-Res after 10 epochs. Here we can see how the increase in adversarial radius by making the model more robust decreases the privacy loss.
\subsection{Background privacy}
The difference between TTM and ATM is drastic when it comes to privacy vulnerabilities. Image data can hide within the adversarial examples but are easier to obtain in ATM. Even with ATM features that are not necessary to identify a particular class remain occluded. Advertisement on trucks and logos on planes can be seen but the specifics cannot be identified and are only seen as blured lines in the reconstructed images.  

\section{Conclusion}
\label{Conclusion}

In this paper we presented the trade-offs between defending against an adversarial attack and a privacy attack by applying three model inversion attacks to three different models. We demonstrated that training a model adversarially (a common adversarial defense) leaves it vulnerable to privacy attacks, namely model inversion attacks that can reconstruct training images directly from the target model. Throughout our three different model inversion attacks, we found it difficult to attack the privacy of TTM while they were trivial to attack adversarially. On the other hand, we were able to better attack ATM but required a greater L2 distance to attack adversarially. 

By looking at the activations of the two models we observed that TTM have adversarial examples with the highest activations of any of the models. The reconstructed images from ATM were found to have a high value in both the TTM and ATM activations. This suggest that these images are minimums for both models if not for the adversarial examples. When these semantically meaningless adversarial examples are removed, as done in the case of an ATM, the highest activations decrease and are left with semantically meaningful minimums. We further analyze this by looking at how these attacks work when seeded with a training image. An ideal model inversion attack would output the same input training image without modification. We observe that the TTM changes the image into an adversarial example while the ATM highlights the key features of the class while blurring out non-key features.

We also observed that when features are not consistent throughout a class' data samples the information is maintained private in the model, such as backgrounds of some classes, but when they are important such as a boat being in the ocean or cars on the road this information is maintained. This is the crucial step in obtaining privacy defenses for ATM. 

The three attacks that we use highlight how traditional model inversion attacks on TTM can be improved with more sophisticated methods but do not provide as much information as a gradient attack on an ATM. The prevalence of adversarial examples in TTM is such that there is almost always an adversarial minimum near any point making model inversion difficult. PGD was only successful with ATM and could only generate a limited amount of images for ATM. Just like for adversarial cases it works well as a standard attack for privacy attacks. DeepDream attacks extend PGD attacks to generate more various images, while maintaining larger structures. DeepDream model inversion attacks improved the results on TTM but still are not capable of providing fine scale reconstructions. When applied on ATM the reconstructions do not improve significantly but are stable enough to receive random noise as input allowing for different images to be generated. 

We also presented a method for quantifying the success of the model inversion metric, by using the output of the last convolutional layers as a feature vector. We then used this feature vector to find the nearest cosine similarity between our model inverted image and the images in the training dataset. The success of the model inversion attack was then calculated as the L2 between the nearest cosine similar image and the model inverted image. 

With ATM increasing the vulnerabilities of privacy attacks, model inversion attacks must be reconsidered when working on private data. Models should at the very least be qualitatively analyzed for data leaks. For the most part, backgrounds are kept private but color and shape of some of the training data is not safe from privacy attacks. More work needs to be done to address the quantification of the success of these privacy attacks, to be able to separate whether an attack extracts a specific training sample or trends in the dataset. Additionally, a better evaluation metric will allow for more advanced attacks and defenses to be developed.


%
%

\bibliography{example_paper}

\begin{thebibliography}{10}

\bibitem{att_faces}
A.~L. Cambridge.
\newblock The at\&t database of faces, 2002.

\bibitem{carlini2017adversarial}
N.~Carlini and D.~Wagner.
\newblock Adversarial examples are not easily detected: Bypassing ten detection
  methods.
\newblock In {\em Proceedings of the 10th ACM Workshop on Artificial
  Intelligence and Security}, pages 3--14. ACM, 2017.

\bibitem{fredrikson2015model}
M.~Fredrikson, S.~Jha, and T.~Ristenpart.
\newblock Model inversion attacks that exploit confidence information and basic
  countermeasures.
\newblock In {\em Proceedings of the 22nd ACM SIGSAC Conference on Computer and
  Communications Security}, pages 1322--1333. ACM, 2015.

\bibitem{goodfellowexplaining}
I.~J. Goodfellow, J.~Shlens, and C.~Szegedy.
\newblock Explaining and harnessing adversarial examples. corr (2015).

\bibitem{hitaj2017deep}
B.~Hitaj, G.~Ateniese, and F.~Perez-Cruz.
\newblock Deep models under the gan: information leakage from collaborative
  deep learning.
\newblock In {\em Proceedings of the 2017 ACM SIGSAC Conference on Computer and
  Communications Security}, pages 603--618. ACM, 2017.

\bibitem{krizhevsky2009learning}
A.~Krizhevsky and G.~Hinton.
\newblock Learning multiple layers of features from tiny images.
\newblock Technical report, Citeseer, 2009.

\bibitem{krizhevsky2010convolutional}
A.~Krizhevsky and G.~Hinton.
\newblock Convolutional deep belief networks on cifar-10.
\newblock {\em Unpublished manuscript}, 40(7), 2010.

\bibitem{liao2018defense}
F.~Liao, M.~Liang, Y.~Dong, T.~Pang, X.~Hu, and J.~Zhu.
\newblock Defense against adversarial attacks using high-level representation
  guided denoiser.
\newblock In {\em Proceedings of the IEEE Conference on Computer Vision and
  Pattern Recognition}, pages 1778--1787, 2018.

\bibitem{madry2017towards}
A.~Madry, A.~Makelov, L.~Schmidt, D.~Tsipras, and A.~Vladu.
\newblock Towards deep learning models resistant to adversarial attacks.
\newblock {\em arXiv preprint arXiv:1706.06083}, 2017.

\bibitem{mahendran2016visualizing}
A.~Mahendran and A.~Vedaldi.
\newblock Visualizing deep convolutional neural networks using natural
  pre-images.
\newblock {\em International Journal of Computer Vision}, 120(3):233--255,
  2016.

\bibitem{mordvintsev2015deepdream}
A.~Mordvintsev, C.~Olah, and M.~Tyka.
\newblock Deepdream-a code example for visualizing neural networks.
\newblock {\em Google Research}, 2:5, 2015.

\bibitem{odena2017conditional}
A.~Odena, C.~Olah, and J.~Shlens.
\newblock Conditional image synthesis with auxiliary classifier gans.
\newblock In {\em Proceedings of the 34th International Conference on Machine
  Learning-Volume 70}, pages 2642--2651. JMLR. org, 2017.

\bibitem{papernot2016distillation}
N.~Papernot, P.~McDaniel, X.~Wu, S.~Jha, and A.~Swami.
\newblock Distillation as a defense to adversarial perturbations against deep
  neural networks.
\newblock In {\em 2016 IEEE Symposium on Security and Privacy (SP)}, pages
  582--597. IEEE, 2016.

\bibitem{salem2018ml}
A.~Salem, Y.~Zhang, M.~Humbert, M.~Fritz, and M.~Backes.
\newblock Ml-leaks: Model and data independent membership inference attacks and
  defenses on machine learning models.
\newblock {\em arXiv preprint arXiv:1806.01246}, 2018.

\bibitem{shokri2017membership}
R.~Shokri, M.~Stronati, C.~Song, and V.~Shmatikov.
\newblock Membership inference attacks against machine learning models.
\newblock In {\em Security and Privacy (SP), 2017 IEEE Symposium on}, pages
  3--18. IEEE, 2017.

\bibitem{simonyan2014very}
K.~Simonyan and A.~Zisserman.
\newblock Very deep convolutional networks for large-scale image recognition.
\newblock {\em arXiv preprint arXiv:1409.1556}, 2014.

\bibitem{tramer2016stealing}
F.~Tram{\`e}r, F.~Zhang, A.~Juels, M.~K. Reiter, and T.~Ristenpart.
\newblock Stealing machine learning models via prediction apis.
\newblock In {\em USENIX Security Symposium}, pages 601--618, 2016.

\bibitem{tsipras2018robustness}
D.~Tsipras, S.~Santurkar, L.~Engstrom, A.~Turner, and A.~Madry.
\newblock Robustness may be at odds with accuracy.
\newblock 2018.

\bibitem{yang2017generative}
C.~Yang, Q.~Wu, H.~Li, and Y.~Chen.
\newblock Generative poisoning attack method against neural networks.
\newblock {\em arXiv preprint arXiv:1703.01340}, 2017.

\bibitem{zagoruyko2016wide}
S.~Zagoruyko and N.~Komodakis.
\newblock Wide residual networks.
\newblock {\em arXiv preprint arXiv:1605.07146}, 2016.

\end{thebibliography}
\bibliographystyle{abbrv}

\end{document}